\pdfoutput=1

\documentclass[11pt]{article}

\usepackage[final]{acl}

\usepackage{amsmath,amsfonts,bm}

\def\eqref#1{equation~\ref{#1}}

\def\1{\bm{1}}

\DeclareMathAlphabet{\mathsfit}{\encodingdefault}{\sfdefault}{m}{sl}
\SetMathAlphabet{\mathsfit}{bold}{\encodingdefault}{\sfdefault}{bx}{n}

\definecolor{darkred}{rgb}{0.8, 0.0, 0.0}
\usepackage{times}
\usepackage{latexsym}
\usepackage{times}
\usepackage{comment}
\usepackage{pifont} 
\usepackage[normalem]{ulem} %
\usepackage{latexsym}
\usepackage{threeparttable}
\usepackage[most]{tcolorbox}
\usepackage{soul}
\usepackage{longtable}
\usepackage{xcolor}
\usepackage{enumitem}
\usepackage[utf8]{inputenc} 
\usepackage[T1]{fontenc}    
\usepackage{hyperref}       
\usepackage{url}            
\usepackage{booktabs}       
\usepackage{amsfonts}       
\usepackage{nicefrac}       
\usepackage{microtype}      
\usepackage{xcolor, colortbl}         
\usepackage{CJKutf8}
\usepackage{graphicx}
\usepackage{float}
\usepackage{amsmath}
\usepackage[ruled]{algorithm2e}
\usepackage{multirow}
\usepackage{array}
\usepackage{algpseudocode}
\usepackage{setspace} 
\usepackage{bbm} 
\usepackage{amsmath}
\usepackage{amssymb}
\usepackage{amsthm}
\usepackage{mathptmx}
\usepackage{subfig}
\usepackage[T1]{fontenc}

\usepackage[utf8]{inputenc}

\usepackage{microtype}

\usepackage{inconsolata}

\usepackage{graphicx}

\title{RAG-Instruct: Boosting LLMs with Diverse Retrieval-Augmented Instructions}

\author{
Wanlong Liu$^{2 \dagger}$,
Junying Chen$^{1 \dagger}$,
Ke Ji$^{1}$,
Li Zhou$^{1}$, Wenyu Chen$^{2}$, Benyou Wang$^{1}$\thanks{Corresponding author. $^\dagger$Equal Contribution.} \\
$^1$ The Chinese University of Hong Kong, Shenzhen, \\
$^2$  University of Electronic Science and Technology of China\\
\textit{wangbenyou@cuhk.edu.cn}\\
}

\definecolor{lightpink}{rgb}{0.945, 0.816, 0.804}
\definecolor{lightgreen}{rgb}{0.851, 0.906, 0.839}
\definecolor{lightblue}{rgb}{0.8, 0.9, 1}
\definecolor{lightyellow}{rgb}{0.992, 0.949, 0.816}

\newtcolorbox[list inside=prompt,auto counter,number within=section]{prompt}[1][]{
    colbacktitle=black!60,
    coltitle=white,
    fontupper=\footnotesize,
    boxsep=5pt,
    breakable,
    enhanced,
    left=0pt,
    right=0pt,
    top=0pt,
    bottom=0pt,
    boxrule=1pt,
    #1   
}
\begin{document}
\maketitle

\begin{abstract}
Retrieval-Augmented Generation (RAG) has emerged as a key paradigm for enhancing large language models (LLMs) by incorporating external knowledge. However, current RAG methods face two limitations: (1) they only cover limited RAG scenarios. (2) They suffer from limited task diversity due to the lack of a general RAG dataset. To address these limitations, we propose \textbf{RAG-Instruct}, a general method for synthesizing diverse and high-quality RAG instruction data based on any source corpus.  Our approach leverages (1) \textit{five RAG paradigms}, which encompass diverse query-document relationships, and  (2) \textit{instruction simulation},  which enhances instruction diversity and quality by utilizing the strengths of existing instruction datasets.  Using this method, we construct a 40K  instruction dataset from Wikipedia, comprehensively covering diverse RAG scenarios and tasks.   Experiments demonstrate that RAG-Instruct effectively enhances LLMs' RAG capabilities, achieving strong zero-shot performance and significantly outperforming various RAG baselines across a diverse set of tasks. RAG-Instruct is publicly available at \href{https://github.com/FreedomIntelligence/RAG-Instruct}{https://github.com/FreedomIntelligence/RAG-Instruct}.
\end{abstract}

\section{Introduction}

Retrieval-Augmented Generation (RAG)~\citep{guu2020retrieval, asai2024reliable} enhances large language models (LLMs) by integrating external knowledge through document retrieval, effectively reducing hallucinations and improving performance across diverse tasks~\citep{asai2023retrieval,jin2024flashrag,lu2022reacc, liu2024compressive}.

Since retrievers are not perfect, and considerable research has shown that noisy retrieval can adversely impact LLM performance~\citep{petronicontext, shi2023large, maekawa2024retrieval}, numerous studies have focused on enhancing the robustness of RAG in handling noisy retrieval contexts ~\citep{wei2024instructrag, chan2024rq}. On the one hand, some studies involve adaptive retrieval based on query analysis~\citep{asaiself, jeong2024adaptive}, or query reformulation~\citep{chan2024rq, ma2023query} to enhance the robustness of LLM-based RAG systems.
On the other hand, ~\citep{zhang2024raft, liu2024chatqa, yoranmaking}  enhance the robustness of models' naive RAG capabilities by training them to adapt to irrelevant and noisy documents.

However, existing RAG methods have two limitations: (1) \textbf{Limited RAG scenarios}. Real-world RAG scenarios are complex: Given the query, the retrieved information may directly contain the answer, offer partial help, or be helpless. Some answers can be obtained from a single document, while others require multi-hop reasoning across multiple documents. Our preliminary study demonstrates existing RAG methods cannot adequately handle all such scenarios~\citep{chan2024rq, asaiself, liu2024chatqa}. (2) \textbf{Limited task diversity}. Due to the lack of a general RAG dataset, most current RAG methods~\citep{wei2024instructrag, zhang2024raft} are fine-tuned on task-specific datasets (e.g., NQ~\citep{kwiatkowski2019natural}, TrivialQA~\citep{joshi2017triviaqa}), which suffer from limited question diversity and data volume.  

To address these limitations, we propose \textbf{RAG-Instruct}, a general method for synthesizing diverse and high-quality RAG instruction data based on any source corpus. Using this method, we construct a 40K synthetic instruction dataset from Wikipedia tailored for RAG.  Our method emphasizes the \textbf{diversity} in two aspects: 
\begin{enumerate}
\item \textbf{Defining diverse RAG paradigms}: we define five RAG query paradigms that encompass various query-document relationships to adapt to different RAG scenarios, considering both document usefulness and the number of useful documents. Based on these modes, we prompt LLMs to synthesize RAG-specific instructions and responses using external documents.

\item  \textbf{Enhancing task diversity and data quality}: we incorporate exemplar data from existing instruction datasets, such as SlimOrca~\citep{mitra2023orca} and Evol Instruct~\citep{Xu2023WizardLMEL}, to guide the generation of RAG instructions. This  approach  is inspired by recent advancements in synthetic instruction datasets which have two key advantages: (1) high-quality 
 instruction-following responses generated by proprietary LLMs, and (2)  diverse instructions that cover a wide range of real-world tasks.  We refer to this approach as ``\textit{Instruction Simulation}'', which leverages the strengths of existing instruction datasets to improve the diversity and quality of the synthesized data. 
\end{enumerate}

Our contributions are summarized as follows:

\begin{itemize} 

\item  We introduce \textbf{RAG-Instruct}, a general method for synthesizing diverse and high-quality RAG instruction data from any given corpus. Using this method, we construct the RAG-Instruct dataset (based on Wikipedia), the first dataset to comprehensively cover diverse RAG scenarios and tasks.

\item  We define five \textit{RAG paradigms} to cover diverse query-document relationships and introduce \textit{Instruction Simulation}, a technique that enhances instruction diversity and quality by utilizing the strengths of existing instruction datasets. These techniques ensure the diversity of synthesized datasets across RAG scenarios and tasks.

\item  Empirical experiments on 11 tasks, including knowledge-intensive QA, multi-step reasoning, and domain-specific benchmarks, demonstrate that RAG-Instruct significantly enhances the model's RAG capabilities.    It significantly outperforms previous state-of-the-art methods such as Self-RAG~\citep{asaiself} and RQ-RAG~\citep{chan2024rq}. Furthermore, ablation studies demonstrate that both \textit{Instruction Simulation} and the five RAG query paradigms significantly contribute to the performance of RAG-Instruct.

\end{itemize}

\begin{table}[htbp]
\centering

\setlength{\tabcolsep}{3pt}
\renewcommand{\arraystretch}{1.1}
\resizebox{\columnwidth}{!}{
\begin{tabular}{@{}lccccc@{}}
\toprule
\multirow{2}{*}{\textbf{Method}} & \multicolumn{3}{c}{\textbf{TriviaQA (Single-hop)}}             & \multicolumn{2}{c}{\textbf{HotpotQA (Multi-hop)}}       \\ \cmidrule(r){2-4} \cmidrule(r){5-6}
                        & Helpful  & Midhelp & Helpless  & Helpful  & Midhelp  
                        \\ \midrule
Llama2-7b               & 71.0             & 48.0               & 17.1             & 51.2             & 21.2               \\
Llama3-8b               & 76.4             & 51.0               & 20.2             & 61.4             & 21.4               \\
Self-RAG (2-7b)    & 77.3             & 42.4               & 14.7             & 45.1             & 16.6               \\
RQ-RAG (2-7b)      & 80.9             & 52.6               & 18.7             & 57.9             & 24.0               \\
ChatQA-1.5 (3-8b)  & 83.5             & 54.9               & 21.4             & 65.1             & 23.9               \\
ChatQA-2.0 (3-8b)  & 82.4             & 51.5               & 20.1             & 61.4             & 19.9               \\
RAG-Instruct (3-8b) & \textbf{86.9}             & \textbf{72.6}               & \textbf{40.5}             & \textbf{73.1}             & \textbf{42.2}               \\ \bottomrule
\end{tabular}}
\caption{Preliminary study of limited RAG scenarios. Accuracy (\%) is reported. We divided TriviaQA and HotPotQA into multiple subsets. More information for each subset is shown in Appendix~\ref{sec:appendix: dividing datasets}}.
\label{tab:exp1}
\vspace{-0.3cm} 
\end{table}

\begin{table}[]

\centering
{
\renewcommand{\arraystretch}{1.1}
\resizebox{\columnwidth}{!}{
\begin{tabular}{lccccccc}
\hline
\multirow{2}{*}{\textbf{Dataset}} & \multirow{2}{*}{\textbf{Data Size}} &  \multicolumn{5}{c}{\textbf{RAG Scenarios}} & \multicolumn{1}{c}{\multirow{2}{*}{\textbf{Task  Diversity}}} \\ \cline{3-7}
                         &                                                        & $r_0$      & $r_1$     & $r_2$     & $r_3$     & $r_4$     & \multicolumn{1}{c}{}                                \\ \hline

LongAlpaca         & 12K                                                  & \ding{55}   & \ding{55} & \ding{55} & \ding{51} & \ding{55}      &  \ding{51}                                                         \\
SQuAD2.0                 &  130K                                 & \ding{55}   & \ding{55} & \ding{55} & \ding{51} & \ding{55}        &   \ding{55}                                                        \\
NarrativeQA              &  15K                                                    & \ding{55}   & \ding{55} & \ding{55} & \ding{51} & \ding{55}        &   \ding{55}                                                        \\
RAG-12000                & 12K                                      & \ding{55}   & \ding{55} & \ding{55} & \ding{51} & \ding{55}      &     \ding{55}                                                     \\ 
Self-RAG Data            &  150K                                                    & \ding{51}   & \ding{55} & \ding{55} & \ding{51} & \ding{55}        &       \ding{55}                                                        \\
RQ-RAG Data              &  40K                                                    & \ding{55}   & \ding{55} & \ding{55} & \ding{51} & \ding{51}       &             \ding{51}                                             \\
\hline
RAG-Instruct     & 40K                                                 & \ding{51}   & \ding{51} & \ding{51} & \ding{51} & \ding{51}        &      \ding{51}                                                    \\ \hline
\end{tabular}}}
\caption{Comparision with existing RAG datasets.$r_0 $ to $r_4$ represent the five RAG scenario paradigms defined in Table~\ref{tab:rag-paradigms}.}
\label{tab:comp}
\end{table}
\section{Preliminary Study}
Since retrievers are not perfect, the helpfulness of retrieved documents to the query varies in real-world scenarios. This raises the question: \textbf{Can existing RAG methods handle complex and various RAG scenarios?} 

To investigate this, we first define five RAG scenarios based on query-document relationships, which we believe cover the majority of RAG use cases: Single-Doc Answer (helpful), Single-Doc Support (midhelp), Useless Doc (helpless), Multi-Doc Answer (helpful), and Multi-Doc Support (midhelp). Detailed definitions for each scenario are provided in \S~\ref{sec:RAG-Instruct}.
\begin{figure*}[ht!]
  \centering
  \vspace{-5pt}
  \resizebox{0.88\textwidth}{!}{
  \includegraphics[width=\textwidth]{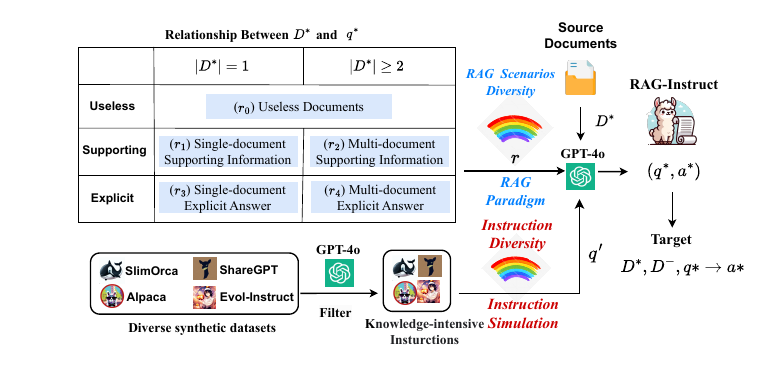}
  }
  \caption{\label{fig:fig31}The main process of synthesizing data with RAG-Instruct. RAG-Instruct ensures instruction data diversity through five RAG paradigms and Instruction Simulation.}
\end{figure*}

Next, we evaluate the performance of existing RAG methods across these five scenarios. Using GPT-4o~\cite{achiam2023gpt}, we categorize questions from two question answering (QA) datasets, Single-hop QA (TriviaQA) and Multi-hop QA (HotPotQA~\citep{yang2018hotpotqa}), into relevant subsets based on the defined RAG scenarios\footnote{We choose these datasets for their large number of questions and subsets, which reduces bias.}. Detailed prompts for categorization are provided in the Appendix~\ref{sec:appendix: dividing datasets}. Then we choose some robust RAG methods, including Self-RAG~\citep{asaiself}, RQ-RAG~\citep{chan2024rq}, ChatQA-1.5 and ChatQA-2.0~\citep{liu2024chatqa} as baselines to explore their performance across the five RAG scenarios. 

As shown in Table~\ref{tab:exp1}, existing RAG methods improve primarily in helpful scenarios, while gains in mid-helpful and helpless scenarios are minimal, with some, such as Self-RAG, even underperforming the baseline. This indicates that existing RAG methods are still unable to handle complex and diverse RAG scenarios effectively. In comparison, our RAG-Instruct method demonstrates significant improvements across all five scenarios, highlighting its effectiveness and adaptability to complex and diverse RAG scenarios.

\paragraph{Comparision with existing RAG datasets.} We review several representative non-task-specific RAG datasets, including long-context instruction datasets such as LongAlpaca~\citep{long-alpaca}, SQuAD2.0~\citep{rajpurkar-etal-2018-know}, and NarrativeQA~\citep{kovcisky2018narrativeqa}, which have been used in ChatQA for RAG training, as well as datasets from classic RAG approaches including Self-RAG data, RQ-RAG data and ChatQA data. As shown in Table~\ref{tab:comp}, existing RAG datasets fail to balance both scenario and task diversity. Long-context instruction datasets are limited to a narrow range of RAG scenarios and focus primarily on reading comprehension tasks.    Additionally, previous state-of-the-art RAG methods such as Self-RAG perform poorly on multi-hop reasoning benchmarks due to their neglect of multi-hop scenarios. 
These shortcomings are reflected in Table~\ref{tab:exp1}.  
In contrast, our RAG-Instruct effectively balances both RAG scenario and task diversity, demonstrating superior generalization and robustness.


\begin{table*}[htbp]

    \centering \small
     \begin{tabular}{>{\raggedright\arraybackslash}m{3.3cm}>{\centering\arraybackslash}m{2cm}c>{\centering\arraybackslash}m{8.5cm}}
        \toprule
       \textbf{$D^*$-$q^*$ Relationship} &  \textbf{Usefulness \newline of $D^*$} & \textbf{$|D^*|$}  & \textbf{Relationship Description} \\ \midrule
       ($r_0$)\newline Useless Doc  & Useless & 1 &   $D^*$ offers no help in answering $q^*$, even if related.\\ \hline
      ($r_1$)\newline Single-Doc Support &  Supporting & 1 &  One doc ($|D^*| = 1$) aids $q*$, providing supporting info or clues without explicit answers.  \\ \hline
       ($r_2$)\newline Multi-Doc Support &  Supporting & $\geq 2$   & Multiple documents ($|D^*| \geq 2$) support $q*$ by providing clues or supporting information without explicitly answering it, requiring integration (multi-hop reasoning).\\ \hline
     ($r_3$)\newline Single-Doc Answer & Explicit & 1  &  One doc ($|D^*| = 1$) directly provides the answer $a^*$ to $q^*$.  \\ \hline
     ($r_4$)\newline Multi-Doc Answer &  Explicit & $\geq 2$  & Multiple docs ($|D^*| \geq 2$) provide a full answer to $q^*$, requiring integration (multi-hop reasoning). \\
        \bottomrule
    \end{tabular}
    \caption{\label{3_1} Descriptions of 5 RAG paradigms. See Appendix~\ref{ap-prompt} for specific prompts.}
    \label{tab:rag-paradigms}
\end{table*}

\section{Method}
This section outlines the RAG-Instruct process, focusing on constructing diverse and high-quality synthetic RAG datasets. The detailed architecture is illustrated in Figure \ref{fig:fig31}.

\subsection{RAG-Instruct}
\label{sec:RAG-Instruct}

\begin{figure*}[ht!]
\centering
\begin{tcolorbox}[colframe=black!50!white, colback=gray!10, sharp corners, fontupper=\small]
\label{fig_prompt}
<Documents> \newline
[1] \textbf{\texttt{\{<document 1>\}}}\newline
[2] \textbf{\texttt{\{<document 2>\}}} \newline
[3] ...\newline
</Documents> \newline

Your task is to generate an English question q* and a corresponding response a* based on the provided <Documents>. Please note that the question q* can take various forms, not limited to questions with a question mark, but also including statements, instructions, and other formats. You need to follow the requirements below to generate the q* and a* \textcolor{blue}{(RAG Paradigms)}:  

\textcolor{blue}{1. The answer to q* can be derived from multiple documents within <Documents>, involving multi-hop reasoning or the integration of information from several documents. \newline
2. a* should leverage the information in <Documents> to provide an accurate answer to q*, ensuring that the response is accurate, detailed, and comprehensive.} 
\newline

Additionally, to ensure diversity, richness, and high quality in the question q* you generate, we will randomly provide a question for you to emulate. In other words, while satisfying the requirements above, make q* similar in task requirement and expression to the \textcolor{darkred}{<Simulated Instruction>} below:  

\textcolor{darkred}{
<Simulated Instruction> \newline
\textbf{\texttt{\{<Simulated Instruction>\}}} \newline
</Simulated Instruction> \newline
}

Please directly generate the question-answer pair (q*, a*) following all the rules above in the format of \{"q*": ..., "a*": ...\}. Ensure the quality of the generated (q*, a*).

\end{tcolorbox}
\caption{The prompt of RAG-Instruct. \textbf{\texttt{{<document>}}} and \textbf{\texttt{{<Simulated Instruction>}}} represent input variables for the document and simulated instruction, respectively. \textcolor{blue}{(Blue text)} indicates RAG Paradigms, illustrating the prompt for $r_4$; other paradigms are shown in Appendix~\ref{ap-prompt}. \textcolor{darkred}{(Red text)} represents Instruction Simulation.}
\label{fig:prompt1}
\end{figure*}

\paragraph{Synthesizing RAG Instructions.} 

Recent proprietary models like GPT-4o~\citep{achiam2023gpt} have demonstrated remarkable capabilities, and many works~\citep{zheng2023pointodyssey, Xu2023WizardLMEL, chen2023huatuogpt} based on synthetic datasets have achieved notable success. Therefore, 
we use GPT-4o to synthesize RAG instructions by leveraging source documents $D^*$\footnote{We will explain how $D^*$ are obtained in the following \textit{Instruction Simulation} section.} to create context-rich instructions.   Specifically, GPT-4o synthesizes an instruction $q^*$ based on $D^*$, followed by a response $a^*$ referencing $D^*$, which can be formalized as:
\begin{equation}
(q^*, a^*) = \mathbf{LLM}(D^*).
\end{equation}

 Inspired by work~\citep{zhang2024raft}, we introduce documents $D^-$ unrelated to $q^*$, which serve as additional noise to enhance the robustness. Then our target RAG instruction is as follows.
\[
    D^*, D^-, q^* \rightarrow a^*.
\]

However, RAG instructions generated this way lack diversity in both RAG scenarios and tasks. To address this, we define five \textbf{RAG paradigms} and introduce \textbf{Instruction Simulation}.

\paragraph{RAG Paradigms.} Real-world RAG
scenarios are complex: Given the $q^*$, $D^*$ may directly contain the answer,
offer partial help, or be helpless. Some answers can be obtained from a single document in $D^*$, while others require multi-hop reasoning across multiple documents. To address this, we define RAG paradigms $\mathbb{R}$, where each $r \in \mathbb{R}$ characterizes the relationship between $D^*$ and $q^*$. As in Table \ref{3_1}, these RAG paradigms consider both document utility and the count of useful documents.

\paragraph{Instruction Simulation.} 

Generating \((q^*, a^*)\) from $D^*$ faces the challenge of instruction monotony. Although $q^*$ is related to $D^*$, the task, phrasing, and difficulty of the instructions can become repetitive with a similar synthesis prompt. Previous datasets address this by broadly collecting instructions~\citep{izacard2023atlas} or using self-instruct~\citep{wang2023self}. In our approach, we leverage diverse, high-quality instructions to diversify $q^*$, a process we term \textit{Instruction Simulation}.

\begin{figure*}[tbp]
    \centering
    \subfloat[\label{fig:RAG dis}Distributions
of RAG Paradigms ]{
    \includegraphics[width=0.39\linewidth]{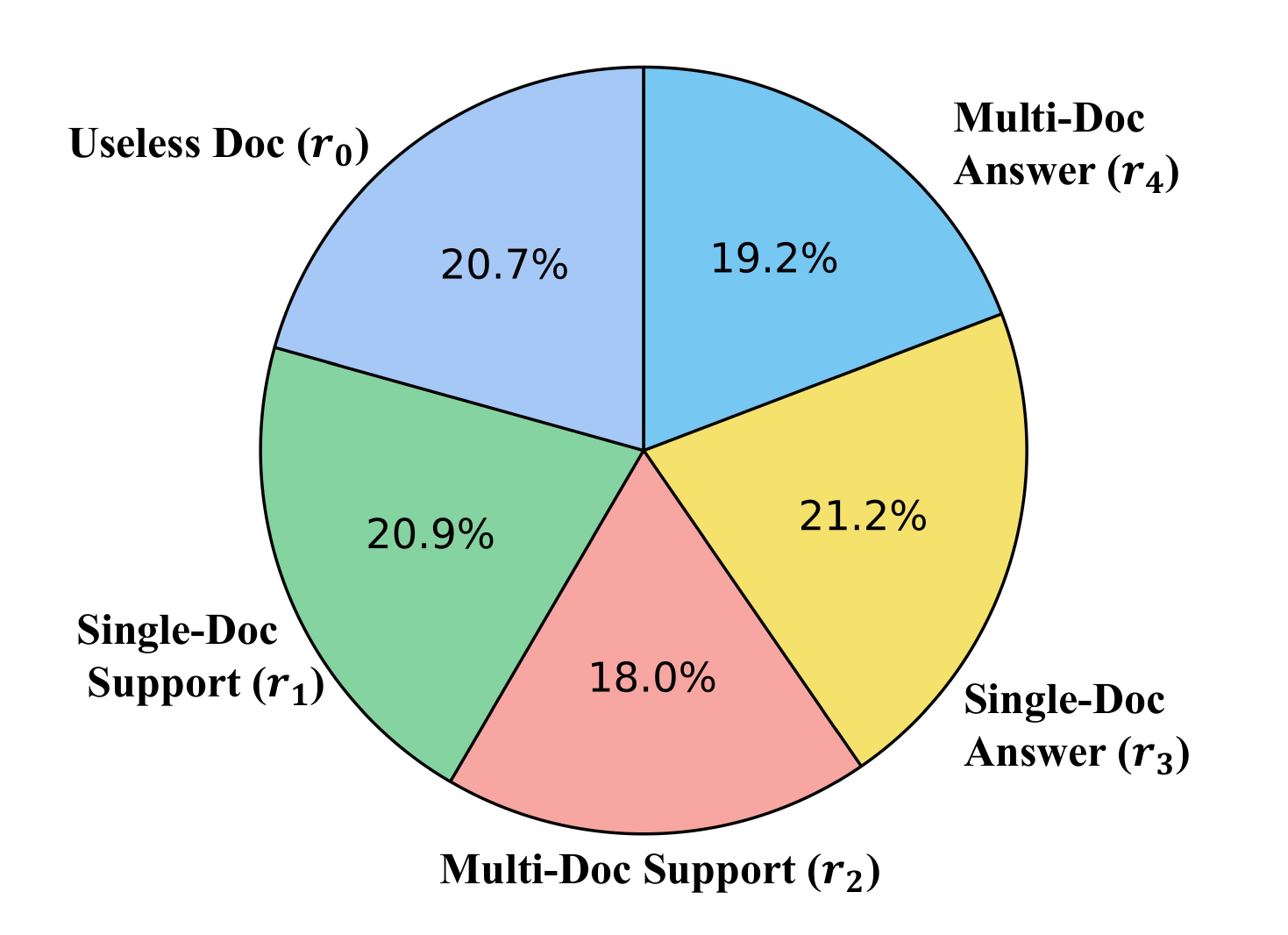}
    }
    \subfloat[\label{fig:data_dis}Distributions
of Data Sources]{
    \includegraphics[width=0.39\linewidth]{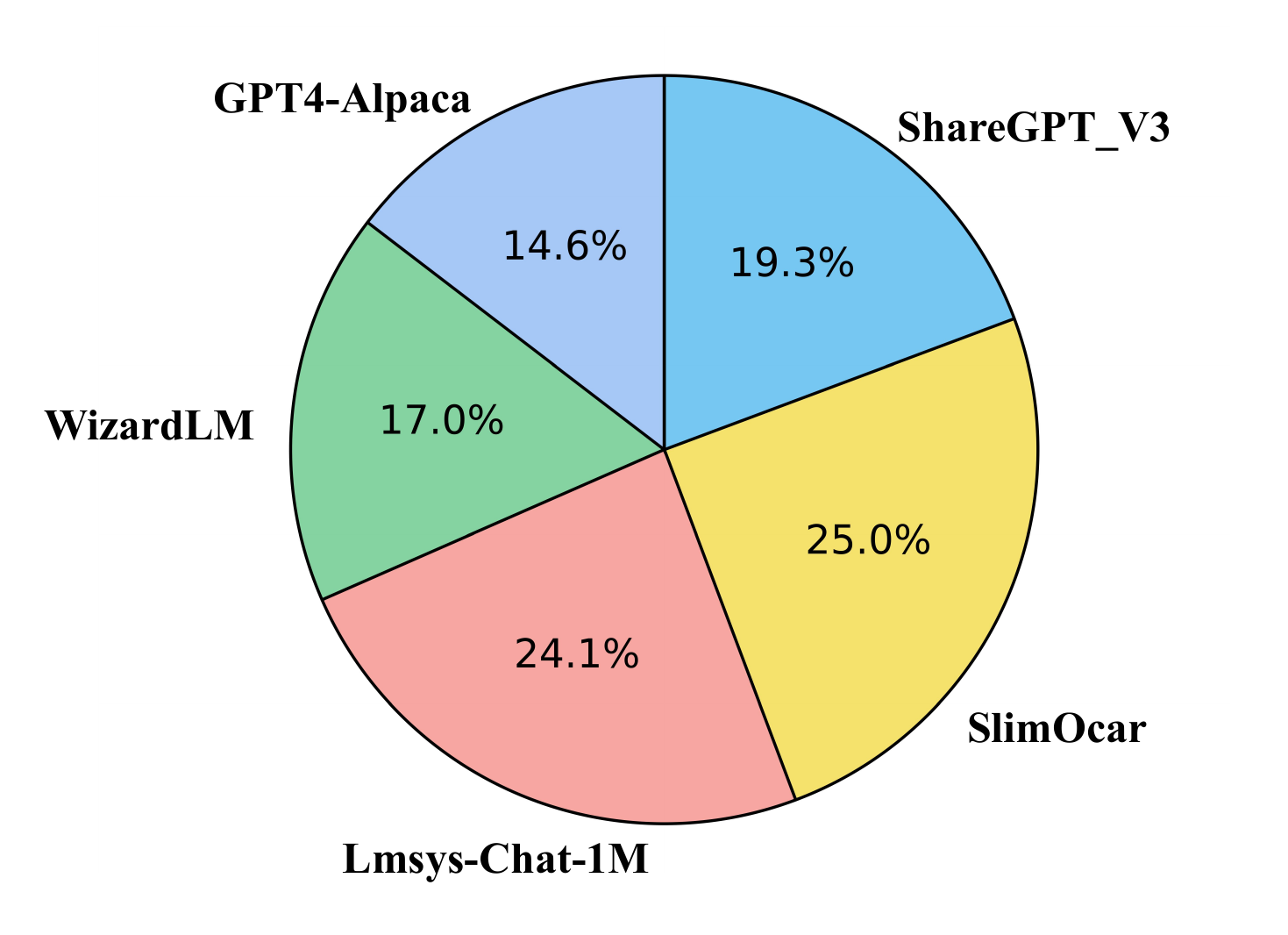}
    }   \caption{The distributions
of RAG paradigms and simulated instruction sources. }
\label{fig:data_distribution}
    
\end{figure*} 
In this process, we use questions from synthetic datasets including ShareGPT~\citep{Wang2023HowFC}, Alpaca~\citep{Cheung2023FactLLaMAOI}, WizardLM-70K~\citep{Xu2023WizardLMEL}, Lmsys-chat-1M~\citep{zheng2023lmsys}, and SlimOrca~\citep{mitra2023orca} as exemplar data. These datasets cover a wide range of tasks, diverse phrasing styles, and varying levels of instruction difficulty. Since RAG is most effective in knowledge-intensive task scenarios~\citep{maekawa2024retrieval, shi2023large}, we use GPT-4o to filter knowledge-intensive instructions from these synthetic datasets (details of the prompt are provided in Appendix~\ref{Sec:train_exp_details}). 

 Then for each synthesis, an instruction $q'\in Q$ is randomly sampled for simulation. Given a corpus $D$ containing multiple documents $d \in D$, the source documents $D^* \subset D$ are retrieved based on  $q'$.  Subsequently, $(q^*, a^*)$ can be synthesized as follows:
\begin{equation}
(q^*, a^*) = \mathbf{LLM}(D^*, q', r),
\end{equation}
where $r$ denotes the sampled RAG paradigm, and the synthesis prompt is illustrated in Figure~\ref{fig:prompt1}. Here, $D^*$ controls the topic of $q^*$, while $q'$ shapes its format and task requirements.

\subsection{Dataset Construction}
We construct RAG-Instruct using Wikipedia corpus. For each synthesis, we sample an RAG paradigm \(r\), a simulated instruction \(q'\), and retrieved  source documents \(D^*\) to generate \((q^*, a^*)\) using GPT-4o.  
To incorporate unrelated documents $D^-$, 
we randomly sample documents retrieved based on $q^*$ and ranked beyond the top 200 as $D^-$. Additionally, for cases where \( |D^*| \geq 2 \), we ensure that the number of source documents is fewer than 5.  Subsequently, \(D^*, D^-, q^* \rightarrow a^*\) is set as the training objective to form RAG-Instruct.  In total, we build a dataset of 40K instructions, with the distributions of RAG paradigms and simulated instructions illustrated in Figure ~\ref{fig:data_distribution}. More dataset construction details are shown in Appendix~\ref{Sec:train_exp_details}.

\section{Experiments}

\subsection{Experimental Settings}
\paragraph{Evaluation Tasks.}
We conduct evaluations of our RAG-Instruct and various baselines across 10 tasks in four major categories: (1) \textbf{Open-Ended Tasks}, including WebQA (WQA)~\citep{berant2013semantic}, PopQA (PQA)~\citep{mallen2023not}, and TriviaQA-unfiltered (TQA)~\citep{joshi2017triviaqa}, where models answer open-domain factual questions with accuracy as the metric. (2) \textbf{Closed-Set Tasks}, including OpenbookQA (OBQA)~\citep{mihaylov2018can}, PubHealth (Pub)~\citep{zhang2023interpretable} and ARC-Challenge (ARC)~\citep{clark2018think}, involving multiple-choice QA with Extract Match (EM) as the metric. (3) \textbf{Multi-Hop Tasks}, including 2WikiMultiHopQA (2WIKI)~\citep{ho2020constructing}, HotpotQA (HotQ)~\citep{yang2018hotpotqa}, and  Musique (MSQ)~\citep{trivedi2022musique}, requiring multi-hop reasoning with accuracy as the metric. (4) \textbf{Domain-Specific Tasks},  CFQA~\citep{chen2022convfinqa} in the financial domain and PubMedQA~\citep{jin2019pubmedqa} in the medical domain, with EM as the metric.   We perform zero-shot evaluations throughout these experiments, providing task instructions without few-shot demonstrations. Reasoning details and prompts are provided in Appendix~\ref{Sec:eval_exp_details}.

\begin{table*}[htbp]

\centering
\resizebox{0.98\textwidth}{!}{\begin{tabular}{lccccccccccc} \toprule
                      & \multicolumn{3}{c}{Open-ended}                & \multicolumn{3}{c}{Closed-set} & \multicolumn{3}{c}{Multi-hop}                 & \multicolumn{2}{c}{Domain-specific} \\ \cmidrule(r){2-4} \cmidrule(r){5-7} \cmidrule(r){8-11} \cmidrule(r){11-12}
                      & WQA         & PQA         & TQA           & OBQA    & Pub       & ARC           & 2WIKI         & HotP        & MSQ       & CFQA             & PubMed           \\  & (acc)         & (acc)         & (acc)           & (EM)    & (EM)        & (EM)           & (acc)         & (acc)        & (acc)        & (EM)              & (EM) \\ \midrule

                       \multicolumn{12}{c}{\it \textcolor{blue}{Closed-Source LLMs} without RAG} \\ 
 GPT-4o  & 68.5 & 60.3 & 79.4 & 88.6 & \textbf{87.7} & \textbf{88.0} & \textbf{88.0} & 54.6 & 31.4 & \textbf{63.0} & 77.0 \\
GPT-4o-mini     & 63.5 & 62.2 & 77.2 & \textbf{89.6} & \underline{87.0} & 84.1 & 74.4 & 54.5 & 30.8 & 60.7 & 73.0 \\
  \midrule
  
\multicolumn{12}{c}{\it \textcolor{blue}{RAG-Specific Models} with RAG} \\ 
RQ-RAG (Llama2-7B)  & 56.5  & 57.1  & 70.2  & 80.6 & 71.8 & 68.3  & 53.7  & 43.1  & 18.2  & 21.9  & 55.6  \\
Self-RAG (Llama2-7B) & 49.0  & 55.8  & 69.3  & 78.0  & 72.4 & 73.1  & 48.4  & 35.8  & 11.5  & 21.5  & 49.8  \\
ChatQA-1.5 (Llama3-8B) & 53.8  & 55.4  & 73.0  & 70.8 & 77.0 & 66.0  & 63.6  & 46.2  & 20.1  & 56.0  & 61.7  \\
ChatQA-2.0 (Llama3-8B) & 50.5  & 58.3  & 72.5  & 72.6 & 75.8 & 65.6  & 59.0  & 42.3  & 16.1  & 51.8  & 61.3  \\
 \midrule

\multicolumn{12}{c}{\it \textcolor{blue}{Open-Source Base Models} with RAG} \\ 
Llama-2-7B  & 49.8 & 51.4 & 62.6 & 56.8 & 36.5 & 48.0 & 55.8 & 38.2 & 17.8 & 22.3 & 58.6 \\
Llama-3-8B  & 59.4 & 57.8 & 71.9 & 58.6 & 50.1 & 50.5 & 62.3 & 42.2 & 23.9 & 44.6 & 62.3   \\ 

Llama-3.1-8B  & 56.7 & 56.8 & 71.5 & 72.4 & 57.6 & 61.4 & 60.7 & 45.5 & 23.5 & 53.1 & 63.0   \\
Qwen2.5-7B  & 61.0 & 58.5 & 71.7 & 70.6 & 56.6 & 65.2 & 59.8 & 46.2 & 22.2 & 52.8 & 67.4   \\

Llama-3.1-70B  & 62.4 & 58.5 & 76.5 & 76.6 & 59.2 & 66.0 & 67.9 & 49.9 & 26.6 & 53.8 & 65.9  \\
\midrule
\multicolumn{12}{c}{\it \textcolor{blue}{Open-Source 
 Instruction-Tuned Models} with RAG} \\
Llama-3-8B-Instruct   & 62.1 & 62.0 & 72.4 & 75.0 & 58.2 & 67.4 & 65.9 & 45.0 & 19.1 & 54.9 & 72.8  \\

Llama-3.1-8B-Instruct   & 61.9  & 62.8  & 73.9  & 77.2  & 56.8 & 70.3  & 66.8  & 45.5  & 19.0  & 53.7  & 73.6  \\
Qwen2.5-7B-Instruct & 64.1  & 62.0  & 75.6  & 74.2  & 74.2 & 75.7  & 66.5  & 49.5  & 20.8  & 58.7  & 62.6  \\

Llama-3.1-70B-Instruct   & 64.9  & 63.3  & 75.4  & 85.0  & 75.4 & \underline{84.7}  & 73.5  & 47.5  & 26.6  & 59.1  & 77.2  \\
\midrule

Llama-2-7B + \colorbox{lightgreen}{RAG-Instruct}  & 67.2  & 62.4  & 77.4  & 71.4  & 75.9 & 74.8  & 68.1  & 53.5  & 21.8 & 29.7  & 71.2  \\
Llama-3-8B + \colorbox{lightgreen}{RAG-Instruct}   & 68.6  & 65.3  & 79.5  & 79.6  & 75.0 & 78.4  & 76.0  & 58.1  & 32.0 & 57.4  & \underline{78.2} \\
Llama-3.1-8B + \colorbox{lightgreen}{RAG-Instruct} & \underline{69.7}  & \underline{68.4}  & \underline{80.0}  & 82.4  & 77.2 & 79.6  & 76.8  & \underline{59.6}  & \underline{33.7} & 57.3  & 77.0 \\ 
Qwen2.5-7B + \colorbox{lightgreen}{RAG-Instruct}  & 66.1  & 63.7  & 78.1  & 78.4  & 76.4 & 78.0  & 74.8  & 54.6  & 27.7 & 55.0  & 72.7 \\
Llama-3.1-70B + \colorbox{lightgreen}{RAG-Instruct}  & \textbf{70.6}  & \textbf{69.4}  & \textbf{82.2}  & \underline{88.6}  & 78.8 & 84.2  & \underline{82.8}  & \textbf{63.9}  & \textbf{41.2} & \underline{61.6} & \textbf{78.5} \\
\bottomrule  
\end{tabular}}
\caption{Zero-shot performance of different instruction datasets on RAG Benchmarks. \textbf{Bold} and \underline{underline} indicate the best and second-best experimental results. The datasets were fine-tuned using identical hyperparameters. }
\label{tab:main_result}
\end{table*}
\paragraph{Baselines.}
We compare our method against a diverse set of baselines, grouped into two main categories: (1) \textbf{Closed-Source LLMs without RAG}, including GPT-4o and GPT-4o-mini. We test them using OpenAI's official APIs. (2) \textbf{Open-source model baselines with RAG}, including Llama2-7b~\citep{touvron2023llama},  Llama3-8b~\cite{dubey2024llama}. Additionally, we also compare with competitive open-source instruction-tuned LMs, such as Llama3-8b-Instruct, Llama3-70B-Instruct, Llama2-7b-chat, Llama-3.1-8B-Instruct and Qwen2.5-7B-Instruct~\cite{yang2024qwen2} to evaluate the advantages of our RAG instruction dataset. For instruction-tuned LMs, we use the official system prompts or instruction formats from their training process when publicly available.  (3) \textbf{RAG-specific baselines}, including Self-RAG, RQ-RAG, ChatQA-1.5, ChatQA-2.0.  For these methods, we evaluate using publicly released model weights and prompts provided by their respective works. 


\paragraph{Training settings.}
We train our model using the RAG-Instruct dataset (wikipedia), which features diverse instruction-following input-output pairs. During the dataset construction, we employ the off-the-shelf Contriever-MS MARCO~\citep{izacardunsupervised} as the retriever. For each data entry, we ensure the use of all source documents \( D^* \) and supplement them with enough unrelated documents \( D^- \) to total 10 documents. For training, we use Llama2-7B, Llama3-8B, Llama3.1-8B, Llama3.1-70B, and Qwen2.5-7B as the base models. Additional training details are provided in Appendix~\ref{Sec:train_exp_details}.

\paragraph{Inference settings.}
We use vLLM~\citep{kwon2023efficient} for memory-efficient inference and adopt a greedy decoding strategy for model generation. For evaluation benchmarks, we utilize Wikipedia as the retrieval corpus and use the Contriever retriever for document retrieval. More detailed inference specifications can be found in Appendix~\ref{Sec:eval_exp_details}.
\begin{figure*}[ht!]
  \centering
  \vspace{-5pt}
  {
  \includegraphics[width=0.99\textwidth]{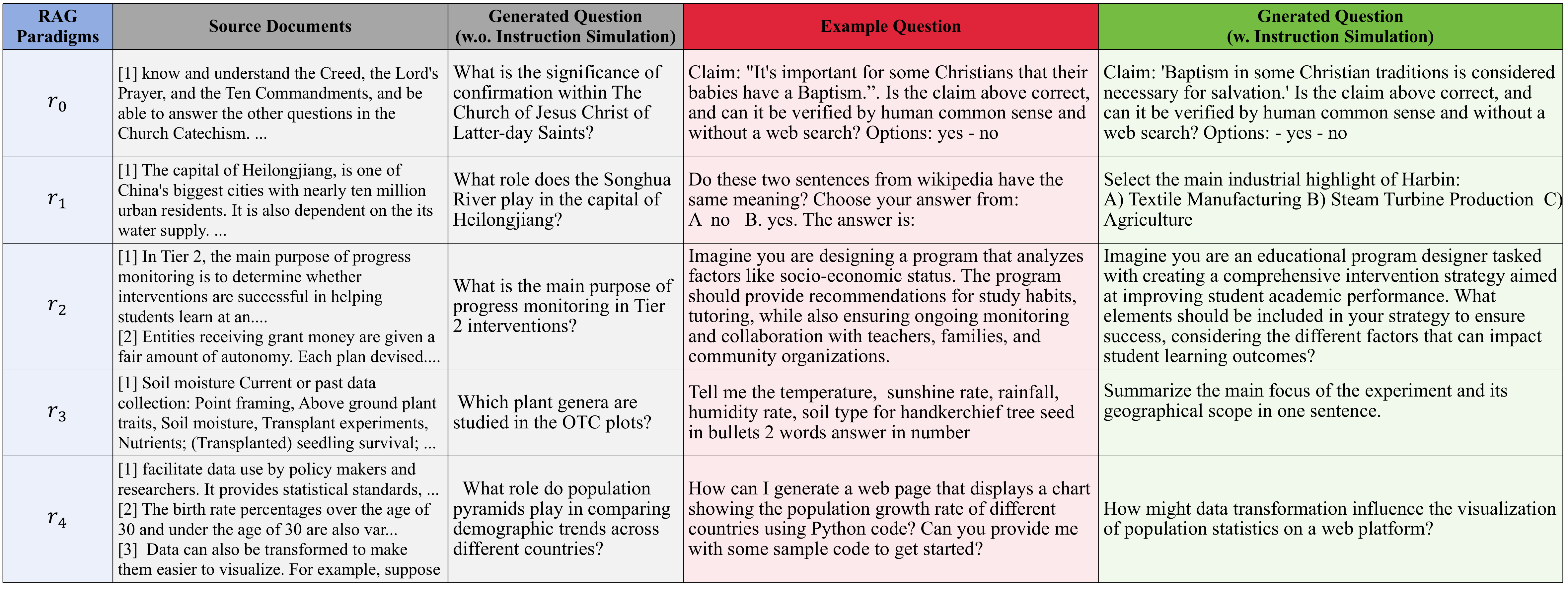}
  }
  \caption{ Some cases of RAG-Instruct for each RAG scenario. We compare the  generated questions with and without using Instruction Simulation. } \label{fig:case}
\end{figure*}
\subsection{RAG Capability Gains}

\paragraph{Comparison against closed-source LLMs.}
As shown in Table~\ref{tab:main_result}, compared to powerful proprietary models like GPT-4o and GPT-4o-mini, our RAG-Instruct, trained on base 8B models, matches or even outperforms them on several tasks, including open-ended tasks (PQA and TQA), multi-hop tasks (HotQA and MSQ), and domain-specific tasks (PubMedQA). This demonstrates that our RAG-Instruct significantly enhances the model's RAG capabilities.

\paragraph{Comparison against RAG-specific models.}
As shown in Table~\ref{tab:main_result}, RAG-specific models such as Self-RAG, and RQ-RAG show significant improvements over the base models on open-ended and closed-set tasks. However, they underperform compared to the base models on domain-specific and multi-hop tasks. In contrast, our RAG-Instruct achieves significant improvements across all four categories of tasks compared to the base models and outperforms all previous SOTA RAG-specific models, particularly in multi-hop and domain-specific tasks. This highlights its superior robustness and generalization across a broader range of RAG scenarios and tasks.

\paragraph{Comparison against Open-source instruction-tuned models.}  We also compare our method with open-source instruction-tuned models, which exhibit strong RAG capabilities. As shown in Table~\ref{tab:main_result}, models trained with RAG-Instruct on base models outperform these instruction-tuned models across various tasks, demonstrating that the RAG instruction dataset effectively enhances the model's RAG performance.

\begin{table*}[htbp] 

\centering
\resizebox{1\textwidth}{!}{\begin{tabular}{lccccccccccc} \toprule
                      & \multicolumn{3}{c}{Open-ended}                & \multicolumn{3}{c}{Closed-set} & \multicolumn{3}{c}{Multi-hop}                 & \multicolumn{2}{c}{Domain-specific} \\ \cmidrule(r){2-4} \cmidrule(r){5-7} \cmidrule(r){8-10} \cmidrule(r){11-12}
                      & WQA         & PQA         & TQA           & OBQA        & Pub   & ARC           & 2WIKI         & HotP        & MSQ       & CFQA             & PubMed           \\ \midrule

$\text{RAG-Instruct}_\text{20k}$ (Llama3-8B)  & 64.6  & 64.8  & 77.0  & 80.2  & 76.0 & 79.4  & 73.0  & 53.1  & 29.7 & 55.4  & 77.2  \\
\quad $\textit{w.o. Simulation}_\text{20k}$ & 63.4  & 63.1  & 75.9  & 74.2 & 71.4  & 70.4  & 62.5  & 47.7  & 25.0  & 47.4  & 70.4  \\


RAG-Instruct $\textit{w.o. Retrieval}$  & 57.6  & 28.4  & 64.2  & 61.2 & 60.6  & 62.8  &47.7  & 35.4  & 10.1  & -  & -  \\
\bottomrule  
\end{tabular}}
\caption{Ablation Study on RAG-Instruct. \textit{w.o. Simulation} indicates the removal of the \textit{Instruction Simulation} process, while \textit{w.o. Retrieval} indicates the performance in non-retrieval scenarios.}
\label{tab:role of instruction simulation}
\end{table*}

\subsection{Impact of Instruction Simulation} 
\label{sec:4.3}
To investigate the impact of \textit{Instruction Simulation}, we design a comparative experiment. We randomly sample a subset \(D_s\) containing 20,000 entries from our RAG-Instruct dataset and create another subset \(D_s'\) without using \textit{Instruction Simulation}. To ensure a fair comparison, \(D_s\) and \(D_s'\) share the same source documents \(D^*\) and include all five RAG scenario paradigms.
We then train two models on Llama3-8B using \(D_s\) and \(D_s'\) with identical hyperparameters. 

As shown in Table~\ref{tab:role of instruction simulation}, removing the \textit{Instruction Simulation} process results in performance declines across all tasks. The drop is smaller for open-ended tasks but significantly larger for closed-set, multi-hop, and domain-specific tasks.   We observe that without \textit{Instruction Simulation}, GPT-4o tends to generate overly simple and uniform questions, resembling open-ended ones, leading to minimal impact on closed-set evaluation. However, the diverse formats of closed-set, multi-hop, and domain-specific tasks, such as multiple-choice and multi-hop reasoning, pose challenges that the model struggles to handle. This highlights the critical role of \textit{Instruction Simulation} in enabling the model to adapt to a wide variety of tasks.

Additionally, we provide specific cases, as shown in Figure~\ref{fig:case}, demonstrating that \textit{Instruction Simulation} generates questions that closely resemble exemplar questions, significantly enhancing diversity compared to those produced without it.  Given the high quality and diversity of the synthesized dataset, \textit{Instruction Simulation} ensures both attributes effectively.

\subsection{Role of RAG Paradigms}

To evaluate the role of RAG paradigms, we design an ablation experiment to verify the effectiveness of the five RAG scenarios in RAG-Instruct. Specifically, we remove the data corresponding to each paradigm from RAG-Instruct one at a time and train models on Llama3-8B using identical training hyperparameters, respectively. 

As shown in Table~\ref{tab:exp_ablation}, when a single RAG paradigm (e.g. $r_0$) is removed from RAG-Instruct, we observe a noticeable performance drop in evaluation benchmarks corresponding to that specific RAG scenario. This indicates that each RAG paradigm plays a critical role in enhancing the model's RAG capabilities across different scenarios. Furthermore, we observe that removing multi-document paradigms ($r_2$ and $r_4$) leads to a significant decline in multi-hop performance. Notably, when all multi-document paradigms ($r_2$ and $r_4$) are removed, the model's performance on multi-hop tasks drops significantly. In contrast, removing all single-document paradigms ($r_0$, $r_1$, $r_3$) results in a relatively small decline in single-hop performance. This suggests that multi-document RAG paradigm data can partially enhance the model's RAG capabilities in single-hop scenarios.

\begin{table}[htbp]
\centering

\setlength{\tabcolsep}{4pt} 
\renewcommand{\arraystretch}{1.3}
\resizebox{\columnwidth}{!}{
\begin{tabular}{@{}llllll@{}}
\toprule
\multirow{2}{*}{\textbf{\Large{Method}}} & \multicolumn{3}{c}{\textbf{\large{TriviaQA (Single)}}}             & \multicolumn{2}{c}{\textbf{\large{HotpotQA (Multi)}}}       \\ \cmidrule(r){2-4} \cmidrule(r){5-6}
                        & \Large{Helpful}  & \Large{Midhelp} & \Large{Helpless}  & \Large{Helpful}  & \Large{Midhelp}  
                        \\ \midrule

\Large{RAG-Instruct} & \Large{86.9}             & \Large{72.6}               & \Large{40.5}             & \Large{73.1}             & \Large{42.2}                       \\ 
\quad \Large{w.o. $r_0$ }              & \Large{86.4}             & \Large{69.6}               & \Large{36.4}{\hspace{0.05cm}\color[HTML]{CD5C5C}{$^{\textbf{–}}$}}           & \Large{74.1}             & \Large{39.3}\\
\quad \Large{w.o. $r_1$}               & \Large{86.5}             & \Large{66.5}{\hspace{0.05cm}\color[HTML]{CD5C5C}{$^{\textbf{–}}$}}                & \Large{40.9}             & \Large{72.4}             & \Large{41.3} \\ 
\quad \Large{w.o. $r_2$}              & \Large{86.2}             & \Large{71.8}               & \Large{39.7}             & \Large{68.2}             & \Large{29.8}{\hspace{0.05cm}\color[HTML]{CD5C5C}{$^{\textbf{–}}$}}  \\ 
\quad \Large{w.o. $r_3$}     & \Large{83.5} {\hspace{0.05cm}\color[HTML]{CD5C5C}{$^{\textbf{–}}$}}              & \Large{70.6}               & \Large{39.6}             & \Large{72.8}             & \Large{42.2}         \\  
\quad \Large{w.o. $r_4$}              & \Large{85.2}             & \Large{72.1}               & \Large{39.5}             & \Large{65.4}{\hspace{0.05cm}\color[HTML]{CD5C5C}{$^{\textbf{–}}$}}              & \Large{38.8} \\

\quad \Large{w.o. $r_0$,$r_1$,$r_3$ }              & \Large{84.3}             & \Large{68.1}{\hspace{0.05cm}\color[HTML]{CD5C5C}{$^{\textbf{–}}$}}               & \Large{36.5}{\hspace{0.05cm}\color[HTML]{CD5C5C}{$^{\textbf{–}}$}}             & \Large{74.8}              & \Large{41.4} \\

\quad \Large{w.o. $r_2$, $r_4$}              & \Large{85.0}             & \Large{71.4}               & \Large{38.8}             & \Large{63.5}{\hspace{0.05cm}\color[HTML]{CD5C5C}{$^{\textbf{–}}$}}              & \Large{26.6}{\hspace{0.05cm}\color[HTML]{CD5C5C}{$^{\textbf{–}}$}}  \\
\bottomrule
\end{tabular}}
\caption{Ablation study on  role of query paradigms. All experiments are conducted based on the Llama3-8B model using identical hyperparameters. `{\color[HTML]{CD5C5C}{${\textbf{–}}$}}' indicates large performance drops for each paradigm. }
\label{tab:exp_ablation}
\vspace{-0.3cm} 
\end{table}
\subsection{Further Analysis}
\paragraph{Performance in non-retrieval scenarios.}

Since our RAG-Instruct is built on the Wikipedia corpus, the performance improvements on evaluation benchmarks may stem from knowledge injection during the supervised fine-tuning stage. To investigate whether our approach genuinely enhances the model's RAG capabilities, we compare the performance in both retrieval and non-retrieval scenarios (based on the Llama3-8B model trained on RAG-Instruct). As shown in Table~\ref{tab:role of instruction simulation}, performance in non-retrieval scenarios is significantly lower across all benchmarks compared to retrieval scenarios, demonstrating that RAG-Instruct effectively enhances the model's capabilities in RAG scenarios.

\paragraph{Different retrieval source.}
To further explore the generalization of our method, we investigate the impact of using different retrieval sources. Specifically, we further evaluate our method on four single-hop QA tasks, including ARC, PQA, TQA and OBQA, utilizing DuckDuckGo, and Bing Search as retrieval sources during inference. The results (detailed in Appendix~\ref{appendix:additionally exps}.) suggest that all retrieval sources effectively improve task performance, with minimal variation in performance across different sources. This demonstrates the robustness of our approach to enhancing RAG capabilities.
\section{Related Work}
Retrieval-augmented generation (RAG) is a widely adopted approach for supplementing the parametric knowledge of large language models (LLMs) with external information sources.   Due to the imperfections of retrievers, the retrieved information often fails to align well with the LLM's needs, which can negatively impact LLM performance~\citep{petronicontext, shi2023large, maekawa2024retrieval}. 

To enhance LLM-based RAG capabilities,  some studies focus on aligning retrievers with LLM needs~\citep{shi2024replug, lin2023ra} through multi-step retrieval processes~\citep{trivedi2023interleaving, jiang2023active, jeong2024adaptive, shao2023enhancing, Yu2023ChainofNoteER, asaiself, wei2024instructrag} and query reformulation~\citep{ma2023query, jeong2024adaptive}. On the other hand, several studies focus on enhancing the RAG capabilities of LLMs by improving their robustness in noisy retrieval contexts. Research such as~\citep{chan2024rq, zhang2024raft, liu2024chatqa, yoranmaking} trains models with additional irrelevant or noisy documents to better handle such scenarios. However, these approaches consider only a limited range of RAG scenarios. Furthermore, the lack of a general RAG dataset forces many works, such as RAFT~\citep{zhang2024raft}, to fine-tune models on task-specific datasets, leading to poor task generalization. This highlights the need for a dataset that covers diverse RAG scenarios and tasks.

\section{Conclusion}

This work introduces RAG-Instruct, a method for synthesizing diverse and high-quality RAG instruction data from any source corpus. It incorporates five RAG paradigms to capture diverse query-document relationships and uses instruction simulation to enhance data quality and diversity by leveraging existing datasets. Using this approach, we construct a 40K instruction dataset from Wikipedia, covering diverse RAG scenarios and tasks.  For future work, we plan to expand the instructions in RAG-Instruct to incorporate chain-of-thought (CoT) characteristics, enabling models to perform planned retrieval based on the query.
\section*{Limitations}
\paragraph{Granularity of RAG Paradigms} While RAG-Instruct introduces five distinct RAG query paradigms to handle various query-document relationships, this relationship is of a coarse granularity. Specifically, the current set of paradigms focuses on broad categories but does not explore more granular or specialized paradigms that could better capture nuanced retrieval tasks. For instance, for multi-hop queries, the number of hops could be specified, and relevance might have more granular options. Expanding the range of RAG paradigms to cover finer distinctions could enhance the model’s ability to handle complex, diverse, and edge-case retrieval situations, thereby improving its robustness and performance.

\paragraph{Reliance on Synthetic Data } Our approach relies on synthetic data generation, which inherently carries the risk of introducing errors or biases, even when using powerful large language models like GPT-4. While the use of large-scale instruction datasets such as SlimOrca and Evol Instruct improves the diversity and quality of the generated data, it is still possible for GPT-4 to produce flawed or inconsistent RAG instructions that may negatively impact downstream tasks. As synthetic data generation becomes more prevalent, ensuring the accuracy and reliability of such data remains an ongoing challenge, especially in high-stakes domains where the correctness of information is critical.

\bibliography{custom}

\clearpage
\appendix
\section{Experimental Details}

\begin{table*}[htbp]
\centering
\resizebox{0.7\textwidth}{!}{
\begin{tabular}{@{}lcccccc@{}}
\toprule
Method & ARC & PQA & OBQA & WQA & AVG.($\uparrow$) & VAR.($\downarrow$) \\ \midrule
       Self-RAG (Llama2-7B)      \\
       \quad + DuckDuckGo & 72.1    & 56.7    & 76.4     & 48.1    &\multirow{3}{*} { 62.9}     &  \multirow{3}{*}{1.9}    \\
      \quad + WIKI & 73.1    & 55.8    & 78.0     &  49.0   &     \\
       \quad + BingSearch & 68.6     &  53.2   & 76.8     & 46.4    &      &      \\ \midrule
      RQ-RAG (Llama2-7B)     \\
       \quad + DuckDuckGo & 69.0    & 58.3    & 79.8    &  52.4   & \multirow{3}{*}{65.2}     & \multirow{3}{*}{1.6}     \\
      \quad + WIKI & 68.3    & 57.1    & 80.6      & 56.5    &      &      \\
       \quad + BingSearch & 68.9    & 55.6    & 78.8     &  57.4   &      &      \\ \midrule
       RAG-Instruct (Llama2-7B)     \\
       \quad + DuckDuckGo & 75.1    &  63.0   & 74.4     & 68.1    & \multirow{3}{*}{\textbf{69.7}}     & \multirow{3}{*}{\textbf{0.7}}     \\
      \quad + WIKI & 74.8    & 62.4    &  71.4    & 67.2    &      &      \\
       \quad + BingSearch & 75.5    & 63.8    & 72.0     & 69.0    &      &      \\ \midrule
\end{tabular}}

\caption{\label{tab:ret_sources}Performance comparison of different retrieval sources. AVG. represents the mean, and VAR. represents the variance.}
\end{table*}

\begin{figure*}[ht!]
\centering
\begin{prompt}[title={Knowledge-Intensive Data Selection Prompt}] 
{
\texttt{\{Question\}}

Please determine if retrieving external information would help answer the above question. If it helps, answer "True", otherwise answer "False".
}
\end{prompt}
\caption{The prompt of filtering knowledge-intensive instructions from synthetic datasets}

\label{fig:prompt2}
\end{figure*}
\subsection{More Details of Training}
\label{Sec:train_exp_details}
\paragraph{Dataset Construction.}
Our RAG-Instruct corpus is built using Wikipedia. Following the approach~\citep {karpukhin2020dense}, each document is a disjoint text block of up to 100 words extracted from a Wikipedia article. Following work~\cite{shi2023large}, we generate Wikipedia document embeddings.

For exemplar data, we select datasets such as ShareGPT~\citep{Wang2023HowFC}, Alpaca~\citep{Cheung2023FactLLaMAOI}, WizardLM-70K~\citep{Xu2023WizardLMEL}, Lmsys-chat-1M~\citep{zheng2023lmsys}, and SlimOrca~\citep{mitra2023orca}. First, we remove overly short, overly long, and low-quality data from these datasets. Then, we randomly sample 120K questions from the filtered data. Since RAG is most effective in knowledge-intensive task scenarios~\citep{maekawa2024retrieval, shi2023large}, we use GPT-4o to further filter for knowledge-intensive instructions from these synthetic datasets. The specific prompt used is shown in Figure~\ref{fig:prompt2}.

\paragraph{Training Details.} We train our models using 8 Nvidia A800 GPUs, each with 80GB of memory. All models are trained for 3 epochs with a total batch size of 128, a peak learning rate of 5e-6, 3\% warmup steps, and linear weight decay. The maximum token length is set to 4096 for all models.
We leverage DeepSpeed Stage 3~\citep{rajbhandari2020zero} for multi-GPU distributed training with BFloat16 precision enabled. FlashAttention~\citep{dao2022flashattention} is employed to improve efficiency during long-context training.

\subsection{More Details of Inference}
\label{Sec:eval_exp_details}
We conduct evaluations of our RAG-Instruct and various baselines across a wide range of downstream tasks, covering 11 tasks in four major categories. Throughout these experiments, we perform zero-shot evaluations, providing task instructions without few-shot demonstrations. For RAG-specific models, we follow the original papers' weights and prompts for inference. For our model and other baselines, reasoning details and prompts are provided in Table~\ref{tab:prompt-template}.

\noindent \textbf{Open-Ended Tasks} include three open-domain question-answering datasets, WebQA (WQA)~\citep{berant2013semantic}, PopQA (PQA)~\citep{mallen2023not}, and TriviaQA-unfiltered (TQA)~\citep{joshi2017triviaqa}, where models are required to answer arbitrary questions based on factual knowledge. We retrieve the top 10 most relevant documents from the corpus as candidate documents. Following ~\citep{asaiself}, we evaluate the performance based on accuracy, assessing whether gold answers are included in the model output.

\noindent \textbf{Closed-Set Tasks} include two multiple-choice question-answering datasets: OpenbookQA (OBQA)~\citep{mihaylov2018can}, PubHealth (Pub)~\citep{zhang2023interpretable} and ARC-Challenge (ARC)~\citep{clark2018think}. We retrieve the top 5 most relevant documents from the corpus as candidate documents. Extract Match (EM) is used as the evaluation metric, and results are reported on the test set for both datasets.

\noindent \textbf{Multi-Hop Tasks} include three multi-hop question-answering datasets: 2WikiMultiHopQA (2WIKI), HotpotQA (HotQ), and  Musique (MSQ). Following~\cite{chan2024rq}, we adopt a reading comprehension setup for these datasets, using candidate documents from their original sources. Each question is linked to 10 passages, with only a few (2 for HotQ and 2 or 4 for 2WIKI) being relevant. MSQ is more challenging, requiring 2, 3, or 4 reasoning hops to answer. We use accuracy as the evaluation metric.

\noindent \textbf{Domain-Specific Tasks} include two datasets: CFQA~\citep{chen2022convfinqa} in the financial domain and PubMedQA~\citep{jin2019pubmedqa} in the medical domain.  For both, we adopt a reading comprehension setup, utilizing the provided context as candidate documents. Exact Match (EM) is used as the evaluation metric.

\begin{table*}[ht]
\centering

\resizebox{1\textwidth}{!}{
\begin{tabular}{ll}
\toprule
\textbf{Task} & \textbf{Template} \\ 
\midrule
\textit{Open-ended} & \begin{tabular}[c]{@{}l@{}} \#\#\# Instruction: \\ Reference Document: \\ \texttt{\{RETRIEVED DOCUMENTS\}} \\ Please refer to the documents above and answer the following question: \\ \texttt{\{QUESTION\}} \\ \#\#\# Response:\end{tabular} \\ \midrule[0.01pt]

\multicolumn{2}{l}{\hspace{-0.22cm}\textit{Domain-specific}} \\ 

\text{OBQA \& ARC} & \begin{tabular}[c]{@{}l@{}} \#\#\# Instruction: \\ Reference Document: \\ \texttt{\{RETRIEVED DOCUMENTS\}} \\ Given four answer candidates, A, B, C and D, choose the best answer choice for the question.\\ Please refer to the documents above and answer the following question: \\ \texttt{\{QUESTION (Including Options) \}} \\ \#\#\# Response:\end{tabular} \\ \midrule[0.01pt]

\text{Pub (FEVER)} & \begin{tabular}[c]{@{}l@{}} \#\#\# Instruction: \\ Reference Document: \\ \texttt{\{RETRIEVED DOCUMENTS\}} \\ Is the following statement correct or not? Say true if it’s correct; otherwise, say false. \\ Please refer to the documents above and answer the following question: \\ \texttt{\{QUESTION\}} \\ \#\#\# Response:\end{tabular} \\ \midrule[0.01pt]

\textit{Multi-hop} & \begin{tabular}[c]{@{}l@{}} \#\#\# Instruction: \\ Reference Document: \\ \texttt{\{RETRIEVED DOCUMENTS\}} \\ Please refer to the documents above and answer the following question: \\ \texttt{\{QUESTION\}} \\ \#\#\# Response:\end{tabular} \\  \midrule[0.01pt]

\multicolumn{2}{l}{\hspace{-0.22cm}\textit{Domain-specific}} \\ 
CFQA & \begin{tabular}[c]{@{}l@{}} \#\#\# Instruction: \\ Reference Document: \\ \texttt{\{RETRIEVED DOCUMENTS\}} \\ Please refer to the documents above and answer the following question: \\ \texttt{\{PREVIOUS QUESTIONS ANSWERS\}} \\ \texttt{\{QUESTION\}} \\ \#\#\# Response:\end{tabular} \\  \midrule[0.01pt] 

PubMed & \begin{tabular}[c]{@{}l@{}} \#\#\# Instruction: \\ Reference Document: \\ \texttt{\{RETRIEVED DOCUMENTS\}} \\ Please refer to the documents above and answer the following question: \\ Answer the question with ``yes'' or ``no'' or ``maybe''. \\ \texttt{\{QUESTION\}} \\ \#\#\# Response:\end{tabular} \\ 

\bottomrule
\end{tabular}}
\caption{{\textbf{Prompt templates} in our Evaluation. For \textit{Open-ended} and \textit{Close-set datasets}, \texttt{RETRIEVED DOCUMENTS} are sourced from the retrieval corpus (e.g., Wikipedia). For \textit{Multi-hop} and \textit{Domain-specific} datasets, \texttt{RETRIEVED DOCUMENTS} come from the context provided in datasets.}} 
\label{tab:prompt-template} 
\end{table*}

\begin{table*}[htbp]
\centering

\resizebox{0.8\textwidth}{!}{
\begin{tabular}{@{}lccccc@{}}
\toprule
\multirow{2}{*}{}                  & \multicolumn{3}{c}{\textbf{TriviaQA(Single-hop QA)}} & \multicolumn{2}{c}{\textbf{HotpotQA (Multi-hop QA)}} \\ \cmidrule(l){2-4}  \cmidrule(l){5-6}
                                   & Helpful        & Midhelpful        & Helpless        & Helpful                 & Midhelpful                 \\ \midrule
\multicolumn{1}{c}{Mumber of Data} & 5628           & 894               & 791             & 4015                    & 3390                       \\ \bottomrule
\end{tabular}}
\caption{Detailed information on dataset subsets categorized into five RAG scenarios.}
\label{tab:data_split_info}
\end{table*}

\section{Detailed Prompts in our Experiments}

\subsection{Prompts for dividing the datasets into five RAG scenarios.}
\label{sec:appendix: dividing datasets}
To explore the performance of RAG methods across five different scenarios, we use GPT-4o to categorize questions from two QA datasets: Single-hop QA (TriviaQA) and Multi-hop QA (HotPotQA), into relevant subsets based on the defined RAG scenarios. The prompts used for categorization are shown in Figure~\ref{fig:divided_single} (Single-hop QA) and Figure~\ref{fig:divide_multi} 
 (Multi-hop QA). The final data volume for each subset is shown in Table~\ref{tab:data_split_info}.

\begin{figure*}[ht!]
\centering
\begin{prompt}[title={Dividing Prompt for Single-hop Question. }] 
{
Documents:\newline
\texttt{\{Doucments\}}  \newline

Question:\newline
\texttt{\{Question\}}  \newline

Answer:\newline
\texttt{\{Answer\}}  \newline

Based on the question and its answer, along with the provided documents, carefully review the documents to assess their overall usefulness in answering the question. Avoid evaluating each document individually; instead, consider the documents as a whole. Choose the most accurate option based on how much the documents contribute to the answer:
1. Very helpful: The answer is directly provided in the documents.
2. Partially helpful: The documents offer supporting information or clues but do not provide an explicit answer.
3. Not helpful: The documents do not contribute to answering the question.
Please directly respond with only the chosen option (1, 2, or 3).
}
\end{prompt}
\caption{The prompt for dividing the single-hop question answering datasets into five RAG scenarios.}

\label{fig:divided_single}
\end{figure*}

\begin{figure*}[ht!]
\centering
\begin{prompt}[title={Dividing Prompt for Multi-hop Question. }] 
{
Documents:\newline
\texttt{\{Doucments\}}  \newline

Question:\newline
\texttt{\{Question\}}  \newline

Answer:\newline
\texttt{\{Answer\}}  \newline

Based on the question and answer provided, carefully review the given documents and assess their overall usefulness in addressing the question. Avoid evaluating each document individually; instead, consider the documents as a whole. Choose the most accurate option based on how much the documents contribute to the answer:
1. Very helpful: The answer can be directly derived from multiple documents.
2. Partially helpful: The documents offer supporting information or clues but do not provide an explicit answer. It needs further reasoning or more knowledge.
Please directly respond with only the chosen option (1, or 2).
}
\end{prompt}
\caption{The prompt for dividing the multi-hop question answering datasets into five RAG scenarios.}

\label{fig:divide_multi}
\end{figure*}

\subsection{
Prompts for synthesizing data for five RAG scenarios.
}
\label{ap-prompt}
We construct five RAG paradigms as described in Figure~\ref{tab:r_0}, Figure~\ref{tab:r_1}, Figure~\ref{tab:r_2}, Figure~\ref{tab:r_3}, and Figure~\ref{tab:r_4}. To generate data for each RAG paradigm, we simply provide the randomly selected source documents \texttt{{<Documents>}} and the simulated instruction \texttt{{<Simulated Instruction>}}.

\section{
Additionally Experiments
}
\label{appendix:additionally exps}
\subsection{Experiments on Different Retrieval Source}
To further explore the generalization of our method, we investigate the impact of using different retrieval sources. Specifically, we further evaluate our method on four single-hop QA tasks, including ARC, PQA, TQA, and OBQA, utilizing DuckDuckGo, Wikipedia, and Bing Search as retrieval sources during inference.  As shown in Table~\ref{tab:ret_sources}, our RAG-Instruct method demonstrates strong resilience to changes in retrieval sources compared to Self-RAG and RQ-RAG. We use the official API to obtain retrieval results.

While Self-RAG, primarily curated using Wikipedia, shows notable performance drops (3-5\%) when switching to Bing Search (with a variance of 1.9), and RQ-RAG similarly experiences performance inconsistencies (variance of 1.6), our RAG-Instruct method exhibits minimal performance fluctuations across different data sources. Specifically, the average performance of RAG-Instruct remains consistently high (69.7) with a variance of only 0.7, even when employing DuckDuckGo, Wikipedia, or Bing Search for retrieval.

This demonstrates that RAG-Instruct not only achieves higher overall performance but also maintains exceptional robustness and stability across diverse retrieval sources, highlighting its superior generalization capabilities compared to existing methods.



\begin{figure*}[ht!]
\centering

\begin{prompt}[title={Useless Doc ($r_0$)}] \label{prompt_r0}
<Documents> \newline
[1] \texttt{\{<Document 1>\}}\newline
</Documents> \newline

Your task is to generate an English question q* and a corresponding response a* based on the provided <Documents>. Please note that the question q* can take various forms, not limited to questions with a question mark, but also including statements, instructions, and other formats. You need to follow the requirements below to generate the q* and a* \textcolor{blue}{(RAG Paradigms)}:  

\textcolor{blue}{1. q* should be related to the <Documents>, but the <Documents> can not provide any useful information for answering q*. \newline
2. a* should be able to answer q*, ensuring that the response a* is accurate, detailed, and comprehensive.} 
\newline

Additionally, to ensure diversity, richness, and high quality in the question q* you generate, we will randomly provide a question for you to emulate. In other words, while satisfying the requirements above, make q* similar in task requirement and expression to the \textcolor{darkred}{<Simulated Instruction>} below:  

\textcolor{darkred}{
<Simulated Instruction> \newline
\texttt{\{<Simulated Instruction>\}} \newline
</Simulated Instruction> \newline
}

Please directly generate the question-answer pair (q*, a*) following all the rules above in the format of \{"q*": ..., "a*": ...\}. Ensure the quality of the generated (q*, a*).
\end{prompt}
\caption{The prompt for synthesizing Useless Doc ($r_0$) data.}
\label{tab:r_0}
\end{figure*}

\begin{figure*}
\begin{prompt}[title={Single-Doc Support ($r_1$)}]\label{prompt_r3}
<Documents> \newline
[1] \texttt{\{<Document 1>\}}\newline
</Documents> \newline

Your task is to generate an English question q* and a corresponding response a* based on the provided <Documents>. Please note that the question q* can take various forms, not limited to questions with a question mark, but also including statements, instructions, and other formats. You need to follow the requirements below to generate the q* and a* \textcolor{blue}{(RAG Paradigms)}:  

\textcolor{blue}{1. <Documents> can support q* by providing useful information or hints, but they do not contain explicit answers.\newline
2. a* should use useful information from <Documents> to aid in answering q*, ensuring that the response is accurate, detailed, and comprehensive.} 
\newline

Additionally, to ensure diversity, richness, and high quality in the question q* you generate, we will randomly provide a question for you to emulate. In other words, while satisfying the requirements above, make q* similar in task requirement and expression to the \textcolor{darkred}{<Simulated Instruction>} below:  

\textcolor{darkred}{
<Simulated Instruction> \newline
\texttt{\{<Simulated Instruction>\}} \newline
</Simulated Instruction> \newline
}

Please directly generate the question-answer pair (q*, a*) following all the rules above in the format of \{"q*": ..., "a*": ...\}. Ensure the quality of the generated (q*, a*).
\end{prompt}
    \caption{The prompt for synthesizing Single-Doc Support ($r_1$) data.}
\label{tab:r_1}
\end{figure*}

\begin{figure*}

\begin{prompt}[title={Multi-Doc Support ($r_2$)}] \label{prompt_r2}
<Documents> \newline
[1] \texttt{\{<Document 1>\}}\newline
[2] \texttt{\{<Document 2>\}} \newline
[3] ...\newline
</Documents> \newline

Your task is to generate an English question q* and a corresponding response a* based on the provided <Documents>. Please note that the question q* can take various forms, not limited to questions with a question mark, but also including statements, instructions, and other formats. You need to follow the requirements below to generate the q* and a* \textcolor{blue}{(RAG Paradigms)}:  

\textcolor{blue}{1. Multiple documents within <Documents> can support q* by providing useful information or hints, but they do not contain explicit answers. \newline
2. a* should use useful information from <Documents> to aid in answering q*, ensuring that the response is accurate, detailed, and comprehensive.} 
\newline

Additionally, to ensure diversity, richness, and high quality in the question q* you generate, we will randomly provide a question for you to emulate. In other words, while satisfying the requirements above, make q* similar in task requirement and expression to the \textcolor{darkred}{<Simulated Instruction>} below:  

\textcolor{darkred}{
<Simulated Instruction> \newline
\texttt{\{<Simulated Instruction>\}} \newline
</Simulated Instruction> \newline
}

Please directly generate the question-answer pair (q*, a*) following all the rules above in the format of \{"q*": ..., "a*": ...\}. Ensure the quality of the generated (q*, a*).
\end{prompt}
     \caption{The prompt for synthesizing Multi-Doc Support ($r_2$) data.}
\label{tab:r_2}
\end{figure*}

\begin{figure*}
 
\begin{prompt}[title={Single-Doc Answer ($r_3$)}]\label{prompt_r3}
<Documents> \newline
[1] \texttt{\{<Document 1>\}}\newline
</Documents> \newline

Your task is to generate an English question q* and a corresponding response a* based on the provided <Documents>. Please note that the question q* can take various forms, not limited to questions with a question mark, but also including statements, instructions, and other formats. You need to follow the requirements below to generate the q* and a* \textcolor{blue}{(RAG Paradigms)}:  

\textcolor{blue}{1. Ensure that q* can be answered directly using the content of <Documents>, meaning its answer can be fully derived from <Documents>. \newline
2. a* should use the information from <Documents> to answer q* accurately, ensuring that the response is accurate, detailed, and comprehensive.} 
\newline

Additionally, to ensure diversity, richness, and high quality in the question q* you generate, we will randomly provide a question for you to emulate. In other words, while satisfying the requirements above, make q* similar in task requirement and expression to the \textcolor{darkred}{<Simulated Instruction>} below:  

\textcolor{darkred}{
<Simulated Instruction> \newline
\texttt{\{<Simulated Instruction>\}} \newline
</Simulated Instruction> \newline
}

Please directly generate the question-answer pair (q*, a*) following all the rules above in the format of \{"q*": ..., "a*": ...\}. Ensure the quality of the generated (q*, a*).
\end{prompt}
    \caption{The prompt for synthesizing Single-Doc Answer ($r_3$) data.}
\label{tab:r_3}
\end{figure*}

\begin{figure*}
 
\begin{prompt}[title={Multi-Doc Answer ($r_4$)}]
\label{prompt_r4}
<Documents> \newline
[1] \texttt{\{<Document 1>\}}\newline
[2] \texttt{\{<Document 2>\}} \newline
[3] ...\newline
</Documents> \newline

Your task is to generate an English question q* and a corresponding response a* based on the provided <Documents>. Please note that the question q* can take various forms, not limited to questions with a question mark, but also including statements, instructions, and other formats. You need to follow the requirements below to generate the q* and a* \textcolor{blue}{(RAG Paradigms)}:  

\textcolor{blue}{1. The answer to q* can be derived from multiple documents within <Documents>, involving multi-hop reasoning or the integration of information from several documents. \newline
2. a* should leverage the information in <Documents> to provide an accurate answer to q*, ensuring that the response is accurate, detailed, and comprehensive.} 
\newline

Additionally, to ensure diversity, richness, and high quality in the question q* you generate, we will randomly provide a question for you to emulate. In other words, while satisfying the requirements above, make q* similar in task requirement and expression to the \textcolor{darkred}{<Simulated Instruction>} below:  

\textcolor{darkred}{
<Simulated Instruction> \newline
\texttt{\{<Simulated Instruction>\}} \newline
</Simulated Instruction> \newline
}

Please directly generate the question-answer pair (q*, a*) following all the rules above in the format of \{"q*": ..., "a*": ...\}. Ensure the quality of the generated (q*, a*).
\end{prompt}
      \caption{The prompt for synthesizing Multi-Doc Answer ($r_4$) data.}
\label{tab:r_4}
\end{figure*}

\end{document}